# Feature selection for medical diagnosis: Evaluation for using a hybrid Stacked-Genetic approach in the diagnosis of heart disease

Jafar Abdollahi [a], Babak Nouri-Moghaddam [a,b,*]

[a] Department of Computer Engineering, Ardabil Branch, Islamic Azad University, Ardabil, Iran
[b] Department of Industrial Engineering, Iran University of Science and Technology, Tehran, 1684613114, Iran

**Graphical Abstract**

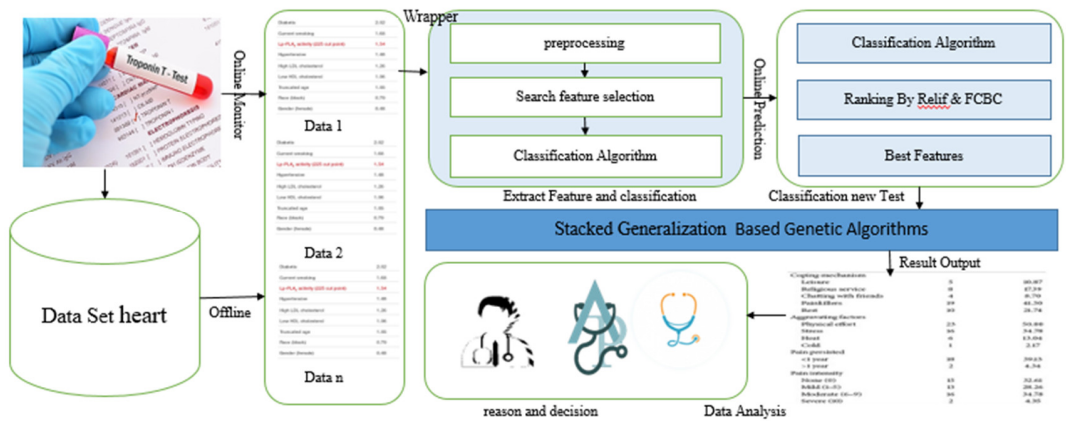

**Abstract**

**Background and purpose**: Heart disease has been one of the most important causes of death in the last 10 years, so the use of classification methods to diagnose and predict heart disease is very important. If this disease is predicted before menstruation, it is possible to prevent high mortality of the disease and provide more accurate and efficient treatment methods.

**Materials and Methods**: Due to the selection of input features, the use of basic algorithms can be very time-consuming. Reducing dimensions or choosing a good subset of features, without risking accuracy, has great importance for basic algorithms for successful use in the region. In this paper, we propose an ensemble-genetic learning method using wrapper feature reduction to select features in disease classification.

**Findings**: The development of a medical diagnosis system based on ensemble learning to predict heart disease provides a more accurate diagnosis than the traditional method and reduces the cost of treatment.

**Conclusion**: The results showed that Thallium Scan and vascular occlusion were the most important features in the diagnosis of heart disease and can distinguish between sick and healthy people with 97.57% accuracy.

1. **Introduction**

Chronic illness is one of the major public health concerns worldwide [1], accounting for more than 50% of global deaths. The most common chronic disease is heart disease, which currently has the highest mortality rate from non-infectious diseases and has a high cost of prevention and treatment [2] Although much research has focused on the importance of coping in managing chronic illness [3]. Heart disease is one of the leading causes of death worldwide, and historically a higher proportion of older people are affected by it, which has changed significantly over the past few years [4-5-6-7]. Traditional methods are not effective in diagnosing such a disease, and it is necessary to develop a medical diagnosis system based on machine learning to more accurately predict and diagnose the disease [8]. Some of the parameters that play an important role in predicting this disease include chest pain, cholesterol, age, blood pressure, and many other factors, and considering these cases, the clinical decision-making system will be improved and also accelerated and will be very useful in predicting heart disease [9].

The heart is responsible for the circulation of blood throughout the body and acts as the body's engine, so heart disease can be fatal. The World Health Organization considers cardiovascular disease to be one of the most important causes of death worldwide [10]. According to the surveys, 56 million people died in 2012 and the most important cause of the mortality rate was heart disease, which is controlled by early detection [11]. The classification of several labels in various applications, including medical diagnosis and semantic annotation, has attracted increasing attention. With such a process, a large number of ensemble approaches have been proposed for multi-tag classification tasks. Most of these approaches build Ensemble members using bag-making schemes, but few have developed a stacked Ensemble approach.

Existing research on stacked Ensemble approaches is still active, but several issues remain (1) Little work has been done to learn the weight of classifications for classification selection. (2) The relationship between paired label correlation and multi-label classification performance has not been sufficiently investigated. To address these issues, we propose an A stacked-genetic approach based on the selection of a new hybrid feature that simultaneously exploits the correlation of tags and the Ensemble members' weight learning process. In our approach, we first develop a weighted stacking Ensemble with a scatter order to facilitate classification selection and construct Ensemble members for multi-tag classification. Second, to improve the classification performance, more paired label correlations are considered to determine the weight of these Ensemble members.

Over the past few years, several studies have been devoted to evaluating the classification accuracy of the various classification algorithms applied to the Cleveland Heard Disease Database [35], which are freely available in the UCI online data mining repository. Due to the need to achieve effective analytical techniques for predicting chronic heart disease, many efforts have been made to improve the quality of evidence-based decisions and recommendations in the information environment. One of the most important functions in health systems is accurate medical advice based on predicting the risk of short-term disease. It is noteworthy that there is a set of disease risk prediction models in the medical literature [12]. Many researchers have endeavored to find the most accurate method of machine learning to explore the relations in heart disease. Given the need, the purpose of this study is to provide an intelligent system for accurate diagnosis and prediction of heart disease and to avoid any unwanted errors, thus reducing medical costs and increasing the quality of treatment [13].

Therefore, in the reference [14], the researchers have presented statistical methods for understanding the three medical data sets to produce some prediction models by extracting appropriate rules to support the decision during the diagnosis process. In this study, methods such as decision tree, NB, SVM, and apriori algorithm have been used which have yielded acceptable results. In Reference [15], a fuzzy system based on a genetic algorithm has been used to predict the risk of heart disease; the results show the high performance of the FDSS approach in disease prediction. In Reference [16], a decision support system has been used to diagnose heart disease in clinical settings. The accuracy of the ANN approach, Cart algorithm, neural network, and logistic regression has reached 97%, 87.6%, 95.6%, and 72%, respectively. In Reference [17], The accuracy of an automated method for early detection of class changes in patients with heart failure using classification algorithms on a data set of 297 patients with evaluation validation approaches has reached 97,87,67%.

Since its inception, this database has been used by many researchers to study different classification problems with different classification algorithms. Detrano in [36] used a logistic regression algorithm and obtained 77.0% classification accuracy. In [37], the author worked on the Cleveland dataset focusing on comparing global evolutionary computational approaches, and observed some improvements in predictive performance when using a new approach. However, the performance of his proposed technique depends on the features selected by the algorithm. Gudadhe et al. [34] realized an architectural base with an MLP network and SVM approach. This architecture achieved an accuracy of 80.41% in terms of classification between the two classes (respectively the presence or absence of heart disease). On the other hand, Homar Kahrananli and Nowruz Allahverdi [38] achieved an accuracy of 87.4% using a combined neural network that is a combination of a fuzzy neural network (FNN) and an artificial neural network

(ANN) [39], And intelligent heart disease prediction system (IHDPS) was introduced. The IHDPS system uses data mining techniques such as DT, NB, and neural networks (NN). Experiments performed by the authors show that the NB model has the best performance in terms of correct prediction (86.12%). The second best NN model with 86.12% correct prediction and the third DT with 80.4% was the score was correct predictions.

Selecting the appropriate features to achieve the best result in data classification has been one of the most challenging topics in recent decades. Although from learning theory, the use of more features increases the accuracy of prediction, practical evidence indicates that this is not always true because not all features are important for detecting the data class label or some of them are irrelevant to the data label. Feature selection methods can be divided into three categories: filtering, packaging, and embedded ones [26-27-28].

1.1 Filtering methods

Filtering methods measure the accuracy of predictions or classifications based on an indirect criterion, such as distance criterion, which indicates how well the classes are separated. Filtering methods are usually used as a preprocessing step. Instead, the features are selected based on their scores on various statistical tests to relate them to the outcome variable [40-47].

1.2 Packaging methods

The packaging method is completely dependent on the classification model and the algorithm determines the optimal subset based on the accuracy obtained from the classifier model and the selection criterion is the same obtained accuracy, and a subset is selected that achieves higher accuracy. In wrapping methods, we try to use a subset of features and teach a model using them [40-47].

1.3 Embedded methods

These methods perform feature selection in the learning process and are usually assigned to a learner. This model also takes advantage of both previous models by using their different evaluation criteria in different search stages. Embedded methods combine filter qualities and packaging methods. This is done by algorithms that have their internal feature selection methods [40-47]. The different techniques of the methods studied are described in Table 1.

**Table 1**
Different techniques of the methods studied

| Filter methods | Wrapper methods | Embedded methods |
| --- | --- | --- |
| information gain | recursive feature elimination | L1 (LASSO) regularization |
| chi-square test | sequential feature selection algorithms | decision tree |
| fisher score | genetic algorithms | |
| correlation coefficient | | |
| variance threshold | | |

1.4 Differences between Filter and Wrapper methods

The main differences between filtering and wrapping methods for selecting features:

- Filter methods are much faster than packaging methods because they do not include model training. On the other hand, packaging methods are also very computationally expensive.
- Filtering methods use statistical methods to evaluate a subset of features, while packaging methods use cross-validation.
- Filtering methods may not find the best feature subset in many cases, but packaging methods can always provide the best feature subset.
- The use of a set of features of packaging methods makes the model more susceptible to the use of a subset of features of filter methods [40-47].

As a result, due to the speed of data collection, the issue of feature selection has become one of the most important issues. In this study, since the use of meta-heuristic algorithms is not enough to solve all the problems because they do not have high speed, so the method of selecting a combination feature (filter, packaging) based on the Ensemble approach has been used. In our current research, our goal is to highlight the comparison of algorithms used in previous studies and their combined use with genetic algorithms and memory construction, so that we can choose the most appropriate prediction method.

We introduce a novel concept, predominant correlation, and propose a fast filter method that can identify relevant features as well as redundancy among relevant features without pairwise correlation analysis. Among the objectives of this article are the following:

- Combining the best Voting machine learning architectures
- Using a hybrid machine learning model to diagnosis of heart disease
- Significant improvement in forecast accuracy
- Use several models in combination
- Achieve a high level of reliability in classification
- Increased accuracy and reduced error compared to single-core models

So the main contribution of this article:

- Machine learning models for heart disease prediction demonstrated high performance.
- Comparing results with the most related researches according to the literature.
- Examining the benefits of ensemble methods proposed recently for prediction.
- using a hybrid Stacked-genetic approach in the diagnosis of heart disease.

The rest of the paper is organized as follows. The proposed model is described in Section 2. In Section 3, experimental results are provided and discussed. Finally, the paper is concluded in Section 4.

## 2. Proposed methodology

We now describe the datasets we chose, the algorithms used, and the experimental methodology. In recent years, the collected data has been used for different research purposes and such dataset may contain thousands of instances (records), each of which may be represented by hundreds or thousands of features (attributes or variables). The large datasets contain a large number of features, many of which contain useful data for understanding information, and a large number of inappropriate and related features. This reduces learning performance and computational efficiency, and a pre-processing step called "feature selection" is used to reduce the dimensions before using any information extraction techniques such as classification, related rules, clustering, and regression [25].

1.2. Dataset

In this study, The data set of heart patients available in (https://archive.ics.uci.edu/ml/datasets/statlog+(heart)) has been used, which has 13 useful variables and 270 records. These variables and abbreviations are listed in Table 2. 75% of the data is for training and 25% of the data is for testing.

**Table 2**
Features of the heart-type dataset

| ID | Meaning | Type |
|---|---|---|
| Age (age in the year) | Patient age | **Integer** |
| sex | Gender | **Numerical two values** |
| chest pain | The location of chest pain | **Numeric values** |
| blood pressure | blood pressure | **Integer** |
| cholesterol | Cholesterol content | **Integer** |
| blood sugar | Blood sugar | **Numerical two values** |
| electrocardiographic | ECG result (electrocardiographic) | **Three-digit number (0, 1, and 2)** |
| heart rate | heartbeat | **Integer** |
| exercise-induced | Angina result from an exercise test | **Numerical two values** |
| depression | ST depression rate | **Real number** |
| slope | ST Exercise Test Result | **Numeric values** |
| ca | Is there a blockage in the arteries or not? | **Numerical two values** |
| Thal | Thalassemia problem | **Numeric values** |
| C(Objective variable) | Is the patient at risk for a heart attack? | **Two-digit number** |

The collection of databases at the University of California, Irvine(UCI) collected by David Aha [12]. The aim of the dataset is to classify the presence or absence of heart disease given the results of various medical tests carried out on a patient. The original dataset contains 13 numeric attributes and a fourteenth attribute indicating whether the patient has a heart condition. This dataset is interesting because it represents real patient data and has been used extensively for testing various data mining techniques. We can use this data together with one or more data mining techniques to help us develop profiles for differentiating individuals with heart disease from those without known heart conditions.

2.3. Data preprocessing

Data preprocessing is necessary to prepare the heart-type data in a manner that a deep learning model can accept. Separating the training and testing datasets ensures that the model learns only from the training data and tests its performance with the testing data. The dataset was divided into training and test data.

3.2. Proposed method

in this paper, we develop an optimization algorithm to achieve the optimal ensemble solution efficiently. Extensive experiments on publicly available datasets and real Cardiovascular and Cerebrovascular Disease datasets demonstrate that our proposed algorithm outperforms related state-of-the-art methods from perspectives of benchmarking and real-world applications.

the Ensemble learning approach (consisting of several machine learning algorithms to achieve a common purpose) based on the hybrid feature selection method has been used to predict heart disease. The proposed method is a combination of statistical analysis methods, machine learning algorithms, and a genetic algorithm, which has the advantages of both filtering and packaging methods to select an optimal subset from the total features space. The performance of the proposed hybrid method is evaluated on the published UCI set and compared with conventional feature selection methods such as relief and FCBF, which are subsets of filtering methods, and the Ensemble learning approach based on the genetic algorithm, which belongs to the Wrapper [13-18-19-20-21-22].

```
Input: Data set D = {(Xi, y1), (X2, y2), ..., (Xm, Ym) };
First-level learning algorithms L1, ..., CT;
Second-level learning algorithm L.
Process:
  for t = 1, ... ,T:
  ht = Lt (D)        % Train a first-level individual learner hy by applying the first-level
                     % learning algorithm Lt to the original data set D
  end;
  D' = θ;            % Generate a new data set
  for i = 1, ... , m:
  for t = 1, ... ,T:
  zit = ht(Xi)       % Use ht to classify the training example Xi
       end;
  D' = D' U{(zil, Zi2, ..., ZiT) ,yi)}
  end;
  h' = L(D').        % Train the second-level learner h' by applying the second-level
   %learning algorithm L to the new data set D'
Output: H (X) = h' (h1 (x), ... ,hT (x))
```

**Fig.1.** Ensemble algorithm

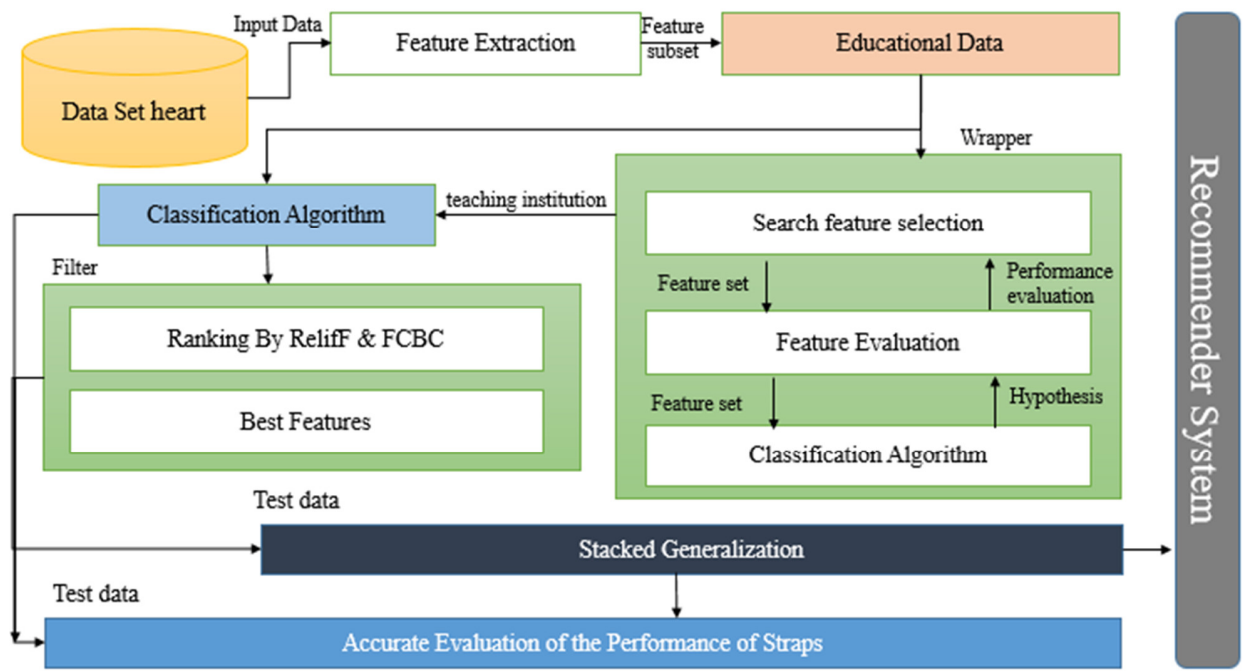

**Fig. 2**. A schematic flow chart of the algorithms.

Feature selection, as a preprocessing step to machine learning, has been effective in reducing dimensionality, removing irrelevant data, increasing learning accuracy, and improving comprehensibility. However, the recent increase of dimensionality of data poses a severe challenge to many existing feature selection methods concerning science and effectiveness. Fig. 2. shows a flow chart of the algorithms we applied in this study. It shows how our hybrid system is designed. We entered the data into our machine learning algorithms in 3 steps. In step one, the basic algorithm was applied individually but in step 2 they were then used in Ensembles. In step 3, we applied our hybrid system comprising of Stacked Generalization and Feather Selection Hybrid (SG-FSH). The algorithms were ultimately evaluated for their performance in predicting the samples.

4.2. Evaluation of Result

For the study, Jupyter notebook was used for implementation and Python programming language was used for coding. We compared the classification performance using Accuracy, Sensitivity, Specificity, Area Under Curve (AUC), Positive predictive value (PPV), Negative predictive value (NPV), F1 score, and Youden Index. FP and FN mean the number of false-positive or false-negative samples.

**Table 3**
Structure of confusion matrix

|  |  | Actual class | |
|---|---|---|---|
|  |  | Negative | Positives |
| **Predicted** | Negative | **True Positives** | False Positives |
| **class** | Positives | False Negatives | **True Negatives** |

TP and TN represent the number of true positive or true negative samples. Specificity measures the ratio of negatives that are correctly discriminated against. Sensitivity measures the ratio of positives that are correctly discriminated against. NPV was used to evaluate the algorithm for screening. PPV was the probability of getting a disease when the diagnostic index is positive. AUC is an index to measure the performance of the classifier. F1 score was a measure of the accuracy of a binary model. Youden Index was the determining exponent of the optimal bound. Additionally, the performance was evaluated with F-measure (F1) to compare the similarity and diversity of performance [21-22].

This article was implemented using the Python programming language version 3.7 in the Anaconda environment of the Jupiter Notebook package on the jupyter notebook platform. Evaluating the accuracy of multi-class algorithms requires complex methods due to the large number of datasets tested and the variety of methods used, as well as the characteristics of the data set, which includes both balanced and unbalanced data. The performance of these algorithms is measured through various parameters such as accuracy, sensitivity, and pressure. Understanding these metrics allows users to understand how well a developed categorization model performs in analyzing textual data. In the field of multi-class problems, traditionally, only the accuracy obtained from the classification is reported as the basic criterion of evaluation and generality [21,22], which is defined as follows: Reciprocal measurement is also described.

**Accuracy**: Indicate the number of "correct predictions made" by the class, divided by the number of "total predictions made" by the same class.

$$Accuracy = \frac{TP+TN}{TP+FP+TN+FN} \tag{1}$$

**Sensitivity**: Real positive rate: If the result is positive for the person, in a few percent of cases, the model will be positive, which is calculated from the following formula.

$$Sensitivity = \frac{TP}{TP+FN} \tag{2}$$

**Properties**: Real negative rate: If the result is negative for the person, in a few percent of cases, the model will also be a negative result, which is calculated from the following formula.

$$Specificity = \frac{TN}{TN+FP} \tag{3}$$

## 3. Experimental results and discussion

Feature selection is one of the most important techniques for reducing dimensions in data preprocessing because datasets generally have redundant and irrelevant features that negatively affect the performance and complexity of classification models. Feature selection has two main purposes, namely to reduce the number of attributes and to increase the classification performance due to its inherent nature [56]. This paper aims to investigate and compare the accuracy of different data mining classification schemes, employing Ensemble Machine Learning Techniques, for the prediction of heart disease. Different classifiers, namely Decision Tree (DT), Naïve Bayes (NB), Multilayer Perceptron (MLP), K-Nearest Neighbor (K-NN), Single Conjunctive Rule Learner (SCRL), Radial Basis Function (RBF), and Support Vector Machine (SVM), have been employed. Moreover, the ensemble prediction of classifiers, bagging, boosting, and stacking, has been applied to the dataset. The results of the experiments indicate that the Stacked-Genetic approach in the diagnosis of heart disease method using the boosting technique outperforms the other aforementioned methods.

Table 4. As described above, filter methods carry out the feature selection process as a pre-processing step with no induction algorithm. The general characteristics of the training data are used to select features. The results obtained for the four filters studied (Relief, FCBF) are compared and discussed. The final aim of this study is to select a filter to construct a hybrid method for feature selection.

**Table.4**
Result of filter method

| Title | FCBF | | Relief | |
|---|---|---|---|---|
| | Running Time | Selected Features | Running Time | Selected Features |
| **heart** | 20 | 8 | 32 | 9 |
| **Accuracy** | 0.872 | | 0.9416 | |

*Note: FCBF=Fast Correlation Based Filter, Relief = filter-method approach to feature selection

Table 5. Seven classifiers were adopted to compare the performance of the selection approaches, using Accuracy and Receiver Operating Characteristics area (aROC) as matric. Result confirms the utility of feature selection for classification and the superiority of wrapper methods. However, some problems do arise from using wrapper methods and, the evidence is proposed that filters are a reasonable alternative with the limited computational cost for dealing with large datasets.

**Table 5**
results of different wrapper method

|  | Accuracy | Sensitivity | Specific |
|---|---|---|---|
| SVM-GA | 84 | 82 | 79 |
| NB-GA | 79 | 81 | 82 |
| Dtree-GA | 94 | 92 | 92.87 |
| MLP-GA | 92 | 91 | 90 |
| KNN-GA | 87 | 86 | 87 |
| RFC-GA | 94 | 93 | 92 |
| LR-GA | 90 | 89 | 91 |
| Stacked-GA | 97.57 | 96 | 97 |

*Note: SVM_GA=Support vector Machine based Genetic Algorithms, NB_GA= Naïve Bayes based Genetic Algorithms, Dtree-GA= Decision Tree Classifier based Genetic Algorithms, MLP_GA= Multi layer perceptron based Genetic Algorithms, KNN_GA= nearest neighbor based Genetic Algorithms, RFC_GA= Random Forest Classifier based Genetic Algorithms, LR_GA= Logistic Regression based Genetic Algorithms and SG_GA= Stacked Generalization based Genetic Algorithms.

In this paper, we intended to use several evolutionary algorithms proposed for feature selection. One of the advantages of the filter, which is a basic component, is that the calculations are obvious, pre-fitting is prevented and it is suitable for certain datasets. However, this method also has disadvantages, including the fact that the unwanted desired subset may be removed from a subset. In general, the main difference between the filtering and packaging methods is that the first one involves a non-repetitive calculation in the database, but the latter can adapt itself to machine learning algorithms and be used. It can be concluded that the packaging results will be better than the filtering method, but the computational cost of this method is high [29-30-31-32]. The results of this article are shown in the table 6.

**Table 6**
Comparison of accuracy rates in Holdout, and Cross-Validation approach. Diagnosis of detection heart disease was compared in rates among various algorithms.

| Model | Holdout | | | K-fold=2 | | | K-fold=5 | | | K-fold=10 | | |
|---|---|---|---|---|---|---|---|---|---|---|---|---|
|  | ACC | Sen | Spec | ACC | Sen | Spec | ACC | Sen | Spec | ACC | Sen | Spec |
| **RF** | 88 | 87 | 86 | 92 | 91 | 90 | 93 | 92 | 91 | 94 | 93 | 92 |
| **KNN** | 76 | 77 | 73 | 77 | 76 | 77 | 81 | 80 | 80.56 | 89 | 85 | 87 |
| **MLP** | 91 | 90 | 92 | 93 | 92 | 91 | 93 | 92 | 91 | 94 | 93 | 92 |
| **Dtree** | 88 | 88 | 87 | 95 | 94 | 96 | 93 | 92 | 89 | 94 | 91 | 93 |
| **NB** | 75 | 74 | 72 | 79 | 78 | 78.5 | 78 | 77 | 76 | 86 | 85 | 83 |
| **LR** | 74 | 73 | 71 | 75 | 74 | 75 | 81 | 80 | 79 | 87 | 85 | 84 |
| **SVM** | 73 | 71 | 70 | 73 | 72 | 69 | 88 | 85 | 84 | 89.05 | 89 | 88 |
| **Suggest Method (ST-GA)** | 88 | 87 | 87.5 | 92.56 | 92 | 93 | 96 | 95 | 96 | 97.57 | 98 | 97 |

*Note: RF= Random Forest, SVM= Support Vector Machine, ANN= Artificial Neural Network, LR= Logistic Regression. GB= Gradient Boosting Classifier, DT= Decision Tree Classifier, NB= Naïve Bayes, Accuracy, Sensitivity, Specificity, PPV= Positive Predictive Value, NPV= Negative Predictive Value, TPR= True Positive Rate, FPR= False Positive Rate.

Increasing advances in computer and electronic technology have provided scientists with the opportunity to collect and study data on various phenomena. Data mining and machine learning are among the important techniques in analyzing and constructing data-driven diagnostic models [57]. The performance of different machine learning algorithms, such as logistic regression, random forest, and support vector machine, etc. was compared with the accuracy based on evolutionary algorithms. The accuracy of the model predictions showed that Ensemble learning based on a genetic algorithm had the best performance and could lead to 97.57% accuracy.

Table 7
Comparison of our results with those of other studies.

| Authors | Method | Classification Accuracy |
|---|---|---|
| Duch et al. [48] | KNN classifier | 85.6 |
| S¸ahan et al. [49] | AWAIS | 82.59 |
| Kahramanli and Allahverdi [50] | Hybrid neural network method | 86.8 |
| Helmy and Rasheed [51] | Algebraic Sigmoid | 85.24 |
| Polat and Gunes¨ ¸ [52] | RBF kernel $F$-score + LS-SVM | 83.70 |
| Karegowda et al. [53] | GA + Naıve Bayes | 85.87 |
| Buscema et al. [54] | TWIST algorithm | 84.14 |
| Tomar and Agarwal [55] | Feature selection-based LSTSVM | 85.59 |
| Chandra Babu Gokulnath1 et al (2018) | SVM-GA | 81.32 |
| Zeinab Arabasadi et al (2017) | NN- Genetic algorithm | 89.04 |
| R. Kannan et al(2018) | RFC method | 80.04 |
| Mehrbakhsh Nilashi et al | (PCA-SVM) | 89.04 |
| Singh, P. P et al (2018). | Neural network method | 88.50 |

## 4. Conclusion

Heart disease is one of the most common chronic diseases and causes of adult death worldwide [33]. Health, Medical Education Department has announced that 33 to 38 percent of deaths in the country are due to cardiovascular disease, and Iran has the highest rate of heart death in the world. Changes in people's lifestyles increase the prevalence of heart disease in Iran. Evidence for lifestyle changes shows that the prevalence of cardiovascular disease is increasing in Iran. It is estimated that by 2020, the mortality caused by these diseases will increase to 25 million. In this paper, an ensemble learning approach based on a genetic algorithm was used to select effective features in the immediate diagnosis of heart disease. The results show the high performance of the proposed method in the immediate diagnosis of heart disease with 97.57% accuracy.

As a suggestion for future research, the Naive Bayesian method, decision tree, or support vector regression can be used as a classifier model and its combination with the new method to select a combination feature extracted from this study can be used to diagnose and predict metastatic diseases of breast cancer, lung cancer, Covid-19 and various diseases.

**Conflict of interest**
The authors declare that no conflict of interest exists
**Acknowledgment**
We are thankful to our colleagues who provided expertise that greatly assisted the research.
**Source of funding**
All the funding of this study was provided by the authors.